\documentclass[nohyperref]{article}

\usepackage{microtype}
\usepackage{graphicx}
\usepackage{subfigure}
\usepackage{booktabs} % for professional tables

\usepackage{hyperref}

\usepackage[accepted]{ai4science} 

\usepackage{amsmath}
\usepackage{amssymb}
\usepackage{mathtools}
\usepackage{amsthm}
\usepackage{todonotes}

\usepackage[capitalize,noabbrev]{cleveref}

\theoremstyle{plain}

\theoremstyle{definition}

\theoremstyle{remark}

\icmltitlerunning{Neural Basis Functions for Accelerating Solutions to high Mach Euler Equations}

\begin{document}

\twocolumn[
\icmltitle{Neural Basis Functions for Accelerating Solutions \\
           to High Mach Euler Equations}



\icmlsetsymbol{equal}{*}

\begin{icmlauthorlist}
\icmlauthor{David Witman}{equal,yyy}
\icmlauthor{Alexander New}{equal,yyy}
\icmlauthor{Hicham Alkandry}{equal,yyy}
\icmlauthor{Honest Mrema}{equal,yyy}
%\icmlauthor{Firstname5 Lastname5}{yyy}
%\icmlauthor{Firstname6 Lastname6}{sch,yyy,comp}
%icmlauthor{Firstname7 Lastname7}{comp}
%\icmlauthor{Firstname8 Lastname8}{sch}
%\icmlauthor{Firstname8 Lastname8}{yyy,comp}
\end{icmlauthorlist}

\icmlaffiliation{yyy}{Johns Hopkins Applied Physics Laboratory, Laurel MD, USA}

\icmlcorrespondingauthor{David Witman}{dwitman1@gmail.com}
\icmlcorrespondingauthor{Alexander New}{Alex.New@jhuapl.edu}
\icmlcorrespondingauthor{Hicham Alkandry}{Hicham.Alkandry@jhuapl.edu}
\icmlcorrespondingauthor{Honest Mrema}{Honest.Mrema@jhuapl.edu}

\icmlkeywords{Machine Learning, Scientific Machine Learning, Neural Networks, Partial Differential Equations, Euler Equations}

\vskip 0.3in
]



\printAffiliationsAndNotice{\icmlEqualContribution} % otherwise use the standard text.

\begin{abstract}
We propose an approach to solving partial differential equations (PDEs) using a set of neural networks which we call Neural Basis Functions (NBF). This NBF framework is a novel variation of the POD DeepONet operator learning approach where we regress a set of neural networks onto a reduced order Proper Orthogonal Decomposition (POD) basis. These networks are then used in combination with a branch network that ingests the parameters of the prescribed PDE to compute a reduced order approximation to the PDE. This approach is applied to the steady state Euler equations for high speed flow conditions (Mach 10-30) where we consider the 2D flow around a cylinder which develops a shock condition. We then use the NBF predictions as initial conditions to a high fidelity Computational Fluid Dynamics (CFD) solver (CFD++) to show faster convergence. Lessons learned for training and implementing this algorithm will be presented as well.
\end{abstract}

% \alex{nothing about the NBF approach relies on the PDEs having discontinuities, right? we're just focusing on those because those are of interest to hypersonics. so can we rephrase the abstract to be about proposing a new approach for solving PDEs in general? and then mention discontinuous euler as the use case}

% \alex{``where solutions can then be used'' -- ``where predictions from the NBF cna then...''?}

% \alex{maybe say that the branch network is the bit that ingests the PDE parameters}

% \alex{probably want to expand POD, CFD, and DeepONet in the abstract the first time they're used}

% \alex{alex test comment}
% \dave{dave test comment}
% \honest{honest test comment}
% \hicham{hicham test comment}

\section{Introduction}\label{sec:intro}

The past five years have seen significant interest in the use of neural networks to understand the behavior of systems governed by differential equations~\cite{Han2018PDEs,Willard2020physicsai,karniadakis2021physics}. For example, methods like physics-informed neural networks~\cite{raissi2019physics,Lu2019deepxde} can solve a given instance of a differential equation or infer its properties given some amount of data and have been applied in domains like electromagnetism~\cite{Khan2022e&m} and blood cell mechanics~\cite{Yazdani2021bloodcell}.

More recently still, the subfield of operator-learning has attained prominence~\cite{lu2019deeponet,li2020fourier,Bhattacharya2021pdes,mao2021deepmmnet}. The objective of operator learning is to learn a data-driven approximation to an operator that defines a differential equation; a trained operator can then produce a solution to this equation at a greatly reduced computational cost. Despite the field's novelty, applications have already been considered in areas like climate science~\cite{Kashinath2021ml_climate}.

Prior to some of these deep network architectures, efforts to reduce the computational cost often attempted to solve a reduced form of the equations of interest. Approaches like the Proper Orthogonal Decomposition (POD)~\cite{willcox2002balanced, gunzburger2007reduced,witman2017reduced,quarteroni2015reduced, berkooz1993proper} use an eigen-decomposition to extract the primary modes of a given solution space. These modes could then be re-projected in the reduced space to significantly reduce the degrees of freedom required to solve the equations. 

The approach being proposed here borrows concepts from the more traditional Reduced Order Modeling (ROM) community and fuses it with the newer deep network architectures to build a Neural Basis Function (NBF) framework for solving differential equations. 
% Additionally, this approach has some flexibility for extensions like partitioning  space to define

% \alex{not sure what is meant by ``partitioning the parametric space'' here}
% \dave{good catch, I never actually finished that thought and looking back I dont think it really fits}

One of the motivating interests of this work will be the focus on how well this approach can resolve solutions with complicated phenomena. The specific phenomenon we will focus on will be solutions with large gradients or discontinuities, as these problems have posed difficulties for both traditional techniques and the more recent Neural Network based approaches. Existing work that considers similar classes of problems includes~\cite{mao2020shockwavepinn} for physics-informed neural networks and~\cite{mao2021deepmmnet} for operator-learning. We first propose the network architecture and training methodology that will be used to learn the underlying basis functions and associated unknown vectors. 

In this paper, we introduce the formalism of the neural basis function (\cref{sec:nbf}) and describe how to apply it to data sampled from a given PDE (\cref{sec:preparation}). In~\cref{sec:euler}, we demonstrate how to apply the NBF approach to the steady-state Euler equations. We evaluate our model in~\cref{sec:nbf_evaluation} with a comparison to vanilla DeepONet in ~\cref{sec:DeepONet} and then show, in~\cref{sec:acceleration}, that predicted solutions from the NBF can be used to accelerate the process of obtaining high-fidelity data from a CFD solver.

\section{Methods}\label{sec:methods}

\subsection{The Neural Basis Function Formulation}\label{sec:nbf}

% \alex{are we just doing a full-on ``unknowns -> branch netowork'' renaming?}
% \dave{Maybe we can define them here and how they are similar to DeepONet}

% \alex{I cut mention of time-dependence from this section since we're only doing stationary Euler, but it's probably worth at least noting we can do this with time-dependent PDEs as well}

The neural basis function framework at its core takes a traditional reduced basis Finite Element Method (FEM)~\cite{quarteroni2015reduced} approach and uses a set of approximating neural networks to learn the underlying bases and explicit unknowns of the governing equations. Our basis-approximation networks are similar to the trunk component of a Deep Operator Net (DeepONet), and our unknowns-approximation networks are similar to the branch component~\cite{lu2019deeponet}, specifically as found in the recent POD DeepONet extension~\cite{lu2022comprehensive}.

% \alex{a thought -- should the paragraph about comparing NBF to POD DeepONet go here? this would let us be more precise in discussing the distinction, and then also gives us a better chance of keeping the ``in this paper'' summary paragraph on the first page.}

We consider a subclass of PDEs defined for a vector field $\mathbf{w}_\psi: \Omega \to R^L$, where $\Omega \subseteq R^M$, that satisfies the following equations:

\begin{equation}
    \mathcal{S}_\psi\mathbf{w}_\psi = \mathbf{0},\,\,\,\,\mathbf{x} \in \Omega
    \label{eq:interior_operator}
\end{equation}
\begin{equation}
    \mathcal{B}_\psi\mathbf{w}_\psi = \mathbf{0},\,\,\,\,\mathbf{x} \in \partial \Omega,
    \label{eq:boundary_operator}
\end{equation}

where $\mathcal{S}_\Psi$ is a (potentially nonlinear) operator involving derivatives with respect to space, $\partial \Omega$ is the boundary of $\Omega$, $\mathcal{B}_\psi$ is a boundary condition operator, and $\psi\in\Psi$ is a set of parameters that fully specifies the PDE and determines its solution. \Cref{sec:euler} illustrates this formulation for a specific problem. In this paper, we consider steady-state problems with no time-dependence, but, similar to physics-informed neural networks~\cite{raissi2019physics}, the NBF approach can be extended to time-dependent problems.
% \alex{does POD need to assume that $\Omega$ is compact or anything?}
% \dave{I don't think so}

In the operator-learning problem~\cite{li2020fourier,lu2019deeponet,Bhattacharya2021pdes}, we learn an operator $\mathcal{F}$ that, for a given novel parameter $\psi$, produces an approximate solution $\hat{\mathbf{w}} = \mathcal{F}(\psi)$ satisfying $||\hat{\mathbf{w}} - \mathbf{w}_\psi||$ under some appropriate norm. This requires a set of training samples of solutions $\mathbf{w}_\psi$ for varying values of $\psi$. 

% \alex{added some text trying to clarify different formulations of the operator-learning problem}

% \alex{if we have space, it could be useful to have some text comparing this to the ``operator-learning'' approach of standard POD, where the $C_{ij}$'s are constants that can be solved for via a newton approach etc. Wanting to better generalize over the $\Psi$ space then motivates formulating $C_{ij}$'s as functions. this might also help explain the use of the term ``unknowns''. we sort of get at this in the following paragraph, but the point could be made more explicitly -- I'm not sure that someone not familiar with POD will know what ``the discrete form of this approach'' is}

To define this vector solution $\hat{\mathbf{w}}$ we introduce a linear combination of reduced basis functions ($\mathbf{\phi}_j(\mathbf{x}) : \Omega \to R^M$) and a set of unknowns ($\mathbf{C}_j(\psi): \Psi \to R^M$), where $j=1,\hdots,n_{BF}$ indexes the set of bases used to define $\hat{\mathbf{w}}$. Specifically,~\cref{eqn:linear_combo} shows how the $i$th component of $\hat{\mathbf{w}}$ is calculated as a linear combination of the basis functions $\mathbf{\phi}_j(\mathbf{x})$ and the unknowns functions $\mathbf{C}_j(\psi)$. The discrete form of this approach has been used successfully as a way to approximate solutions, within the parameters space $\psi$, of PDEs at a fraction of the computational cost~\cite{willcox2002balanced, gunzburger2007reduced, witman2017reduced}.

% \alex{maybe we can add a comment here that's similar to the discussion at the RQ brownbag, in that POD lets you interpolate through a parameter space of PDEs, given some existing high-fidelity solutions?}

\begin{equation}
    \hat{w}_i(\mathbf{x},\psi) = \sum_j^{n_{BF}} C_{ij}(\psi)\phi_{ij}(\mathbf{x})
    \label{eqn:linear_combo}
\end{equation}

Basis functions are trained by solving the data-driven minimization problem:

\begin{equation}
    \min_{\phi_{i,j}}\sum_l |\phi_{i,j}(\mathbf{x}_l) - U_{i,j}(\mathbf{x}_l)|^2,
    \label{eqn:basis_min}
\end{equation}

% \dave{I replaced $w_i$ with $U_i$ since that is a little more consistent with what we describe in the next section. Also $w_i$ seems to imply we are regressing to the state variables}
% \alex{that's a good point}

where the ground truth basis data points $U_{i,j}(\mathbf{x}_l)$, with $l=1,...,n$ representing the index of all the spatial data-points. We obtain the truth basis data via a process described in \cref{sec:preparation}. 

Unknowns functions are trained by solving the following physics-informed minimization problem:

\begin{equation}
    \min_{C_{i,j}} \sum_{\psi \sim \mathbb{P}_\psi}\sum_{\mathbf{x}\sim\mathbb{P}_\mathbf{x}}||\mathcal{S}_\psi\hat{\mathbf{w}}(\mathbf{x},\psi)]||^2 + ||\mathcal{B}_\psi\hat{\mathbf{w}}(\mathbf{x},\psi)]||^2,
    \label{eqn:unknowns_min}
\end{equation}
where $\mathbb{P}_\psi$ and $\mathbb{P}_{\mathbf{x}}$ are probability distributions defined for the PDE parameters $\psi$ and the spatial inputs $\mathbf{x}$, respectively.

% \alex{TODO: this doesn't include the boundary condition error}

% \alex{TODO: this doesn't talk about the pretraining procedure}

Due to the presence of $\mathcal{S}_\psi$ in \cref{eqn:unknowns_min}, evaluating the loss function will require taking derivatives of $\hat{\mathbf{w}}$ with respect to $\mathbf{x}$. This can be done using modern automatic differentiation techniques~\cite{Baydin2017autodiff}. In practice, we implement our networks in \texttt{PyTorch}~\cite{pytorch} and use the \texttt{functorch}~\cite{functorch} library to calculate loss function derivatives. %Code implementing our method will be made public upon publication.%\footnote{placeholder link to a github repo}.

% \alex{can we write out what minimization problem $\tilde{C}_{i,j}$ is the solution to?}
 
% \alex{this doesn't mention the quadrature weights}
% \dave{I think that is ok for now, tbh for the euler equations I dont think we really need the weak form since we only have first order derivs}

% \vspace{1cm}

% % \alex{what we don't convey here is that the physic-informed loss is regularizing the solution to conform with a known governing equation}
% % \dave{TODO: Yes we need to add this as an additional loss term}

% \vspace{1cm}

% \alex{we can probably hit more that the loss doesn't care about our snapshot data}
% \dave{Well, we need both right? Optimally, we wouldnt need any data to train the unknowns but practically, that doesnt seem to be working.}

% \vspace{1cm}

% \alex{we could maybe note here that the method doesn't require $\Omega$ be a uniform grid or whatever, since that's an assumption of something like the FNO}

\subsection{Data-driven basis construction}\label{sec:preparation}

% \alex{trimming some text here, as it was redundant given previous sections}

Training the basis functions~\cref{eqn:basis_min} requires data, which we obtain by using a high-fidelity simulator like CFD++ to solve instantiations of the PDE. We sweep a subset of our parameter space $\Psi$ and generate field values $\mathbf{w}$ for each sampled $\psi\in\Psi$. These samples are divided into training, validation, and test sets as normal in ML.

For training, we will refer to the total number of training instances as $D$, and we let $\psi_d$ refer to the parameters for the $d$th training instance. We assume that all of our snapshots $\{(\mathbf{w}_d, \psi_d):d=1,\hdots,D\}$ are evaluated at the same space points. We index these points as $\{\mathbf{x}_l\}_{l=1}^n$. Thus, the training set is specified by $\{(\{\mathbf{w}_d(\mathbf{x}_l)\}_{l=1}^n, \psi_d)\}_{d=1}^D$

% \alex{maybe worth noting (1) what the rough dimensionality of $\psi$-space is or cam be, and (2) if it is likely to have any structure / if each component is more or less independently sampled}
% \dave{(1) dimensionality of the parameter space? I mean it could be as large as you wanted right assuming you have enough data (2) Not sure what you mean here?}
% \alex{so when we're sampling points $\psi$, if $\psi$ is a vector, each of its components might be independent (like for the burger's equation), or maybe there's some correlation -- making up something, mach number and angle of attack are positively correlated. this is probably too waffley to matter here though}

% \alex{adding text about how the snapshot datapoints are all evaluated at the same space-time points}

Once the snapshot data is collected, we follow the Snapshot Proper Orthogonal Decomposition method~\cite{sirovich1987turbulence} and perform singular value decompositions (SVDs) on the data to extract their bases. For each component $w_i$ of $\mathbf{w}$, we construct a matrix $\mathcal{W}_i$ containing the observed data points, and then we decompose it as $\mathcal{W}_i = U_i \Sigma_i V_i^T$. 
% \dave{TODO: Figure out consistent notation for points in x}
\begin{equation}
U_i\Sigma _i V_i^T = \mathcal{W}_i = 
\begin{bmatrix}
w_i(\mathbf{x}_1|\psi_1) & \hdots & w_i(\mathbf{x}_1|\psi_D)\\
w_i(\mathbf{x}_2|\psi_1) & \hdots & w_i(\mathbf{x}_2|\psi_D)\\
\vdots & \ddots & \vdots \\
w_i(\mathbf{x}_n|\psi_1) & \hdots & w_i(\mathbf{x}_n|\psi_D)
\end{bmatrix}
\label{eq:decomposition}
\end{equation}
The $U_i$ matrix captures the row-space of the snapshot set and represents the primary space modes of the set of solutions.

Once we have generated the set of decomposed spatial modes describing the parametric solution space, we can generate a bound on the expected reconstructed error levels. A well known POD error bound~\cite{witman2017reduced} can be derived from the singular values
% \alex{an error bound for what? the reconstruction error?} 
in the decomposition via the equation $\sum _{d=n_{BF}+1} ^D \sigma _d ^2$ where $D$ here represents the total number of parameter snapshots available. This error bound allows use to a-priori choose a suitable error tolerance before initiating the full training process. For any of the results presented here, we consider the total number of basis functions as equal to the number of training scenarios.
% \alex{what is meant by "limiting dimensionality" here?}
% \dave{fixed it, the only case where $D$ wouldnt be the number of snapshots is an edge case where the number of points in $\mathbf{x}$ is really small}
% In general the number of solution snapshots $d$ in our parameter space is significantly less than the degrees of freedom (columns of snapshot matrix) so $D$ here is the same as $\Psi_D$.
% \alex{the point of this is that we're able to choose the basis dimension we want for a given error tolerance, right? if so, can we be more explicit about that?}

\subsection{Euler Equations}\label{sec:euler}

\begin{figure}
    \centering
    \includegraphics[width=\linewidth]{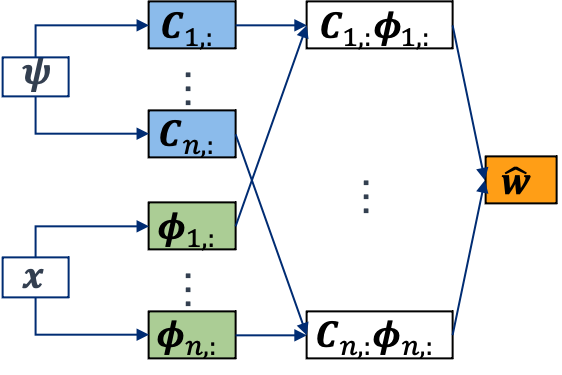}
    \caption{The NBF formulation ingests spatial points $\mathbf{x}$ and PDE parameters $\psi$ into separate sets of networks $\mathbf{\phi}$ and $\mathbf{C}$, and then the predicted state $\hat{\mathbf{w}}$ is a linear combination of these terms.}
    \label{fig:nbf_schematic}
\end{figure}

We consider the steady-state compressible Euler equations~\cite{certik2017theoretical}, defined for a state vector $\mathbf{w}_{\psi} = [\rho, u, v, E]^T$ on a domain $\Omega \subseteq R^2$:

% \alex{not sure if it makes sense to talk about the geometry of $\Omega$ here or in a later section}

\begin{equation}
    \mathcal{S}_\psi \mathbf{w}_{\psi} := \partial_x \mathbf{F}_1(\mathbf{w}_{\psi}) + \partial_y \mathbf{F}_2(\mathbf{w}_{\psi}) = 0
    \label{eq:euler}
\end{equation}

where 

\begin{equation}
\mathbf{F}_1 = \begin{bmatrix}
    \rho u \\
    \rho u^2 + p\\
    \rho vu\\
    (E + p)u
\end{bmatrix},\,\,\,\,\mathbf{F}_2 = \begin{bmatrix}
    \rho v \\
    \rho uv\\
    \rho v^2 + p\\
    (E + p)v
\end{bmatrix}.
\label{eq:euler_components}
\end{equation}

The components of the state vector $\mathbf{w}_{\psi}$ include the fluid density $w_0 = \rho$, the $x$ and $y$ velocities $w_1 = u$ and $w_2 = v$, and the total energy $w_3 = E$. 

The Euler equations \cref{eq:euler} also include the system pressure $P$, which is determined by a closure relation in terms of the other state variables:

\begin{eqnarray*}
        P &=& (\gamma - 1)[E - \frac{1}{2}\rho(u^2 + v^2)]\\
          &=& (\gamma - 1)[w_3 - \frac{1}{2}w_0(w_1^2 + w_2^2)]
\end{eqnarray*}
where $\gamma$ is the gas constant and is set to 1.4, using the perfect gas assumption.

Ultimately, we are interested in using solutions to these equations as initial conditions for a high fidelity CFD solvers to accelerate convergence. In additional term the state variable terms we will also need a relation for the temperature, $T$, which can be expressed as:  

\begin{equation}
    T = \frac{P}{R\rho} = \frac{P}{R w_0}
\end{equation}
where $R$ represents the ratio of specific heats and for this problem we set it to a constant $287.058$, also under the perfect gas assumption. The final term that will be referenced in this example is the speed of the flow and accounts for the magnitude of the $x$ and $y$ components of velocity ($|\mathbf{u}| = \sqrt{u^2 + v^2}$). With these variables we can write the equation for the Mach number:

\begin{equation}
    M = \frac{|\mathbf{u}|}{\sqrt{\gamma R T}}
    \label{eqn:mach_number}
\end{equation}
which will be used in the inlet boundary condition described in \cref{sec:nbf_evaluation}. For further expansion of the Euler equations in terms of our state variables please see \cref{sec:appendix_euler_expansion}.

\subsection{Related work}\label{sec:related-work}

The operator-learning problem is frequently viewed as learning mappings between function-spaces~\cite{lu2019deeponet,li2020fourier,Bhattacharya2021pdes}, where the input function might be an initial condition or forcing function. We consider a more restricted setting, in that we require our operators $\mathcal{S}_\phi$~\cref{eq:interior_operator} and $\mathcal{B}_\psi$~\cref{eq:boundary_operator} to be parametric functions of a finite-dimensional vector $\psi$. In both cases, however, the output is a function~\cref{eqn:linear_combo} defined over an entire domain.

Of operator-learning approaches, the NBF is most similar to the DeepONet~\cite{lu2019deeponet}. Both use multiple types of networks that ingest different types of data (called ``branch'' and ``trunk'' networks in~\cite{lu2019deeponet}). In particular, the NBF can be viewed as a extension of the recently-proposed POD DeepONet~\cite{lu2022comprehensive}. The POD DeepONet retains the full discrete set of bases~\cref{eq:decomposition}, but the NBF regresses these bases onto a set of neural networks $\phi_{i,j}$. This has a few benefits: It 1) can interpolate the modes to new regions of space when data are only available on sparse or unstructured grids; 2) is useful for lower memory applications where storing the potentially large full reduced basis is not feasible; and 3) allows for the ability to compute spatial and/or temporal derivatives of modeled quantities like fluid velocity in a batch process via GPU or another distributed architecture. 

The NBF formulation, like the DeepONet formulation, does not impose any structure to the domain $\Omega$. This is in contrast to some methods like the Fourier Neural Operator~\cite{li2020fourier}. Our use case for the Euler equations in~\cref{sec:nbf_evaluation} relies on this flexibility. See also~\cite{lu2022comprehensive} for a discussion of how different operator-learning approaches deal with non-uniform domains.

\section{Results}\label{sec:results}

\subsection{Evaluation of NBF for Euler Equation}\label{sec:nbf_evaluation}

\begin{figure}[h]
    \centering
    \includegraphics[width=.7\linewidth]{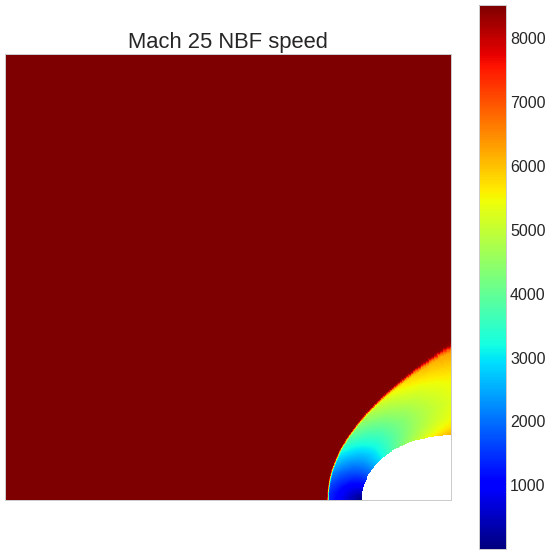}
    \caption{The predicted NBF approximation of speed $\sqrt{u^2+v^2}$ at the test point, Mach 25. The majority of the domain (solid color) has high values of speed, then, at the shockwave (bottom right), the speed distribution rapidly changes.}
    \label{fig:nbf_mach20_speed}
\end{figure}

For this work, we will make use of a rectangular domain with a quarter circle cutout located at $x=[0,0]$ with radius 1 (see \cref{fig:nbf_mach20_speed}). There is an inflow condition on the left side of the domain which is controlled by the Mach number, which for this study is the sole parameter in $\psi$. The  domain extends to $-4$ in the $x$-direction and $7$ in the $y$-direction. This domain was chosen to allow sufficient room for the shock condition based on our predefined parametric Mach number range. This 2D cylinder in cross-flow problem is often used as a benchmark in the CFD space for testing different algorithms and their ability to capture shockwaves. 

We then generated a steady-state dataset of high-fidelity solutions ranging from Mach 10 to Mach 30 in increments of 1. This data was generated using CFD++ in a time-dependent configuration and allowed to evolve until the time derivative portion of Euler equation sufficiently decreased. CFD++ is an unstructured finite volume flow solver that is maintained by Metacomp Technologies Inc. It has the capability of solving the full Navier-Stokes equations with high-order spatial and temporal accuracy. To resolve the inviscid Euler equations, the solver uses a pseudo time-marching approach where the initial condition is integrated over time until convergence is reached. The convergence can be accelerated by using spatially varying time steps. CFD++ version 20.1 was utilized to perform all the simulations.

Once we have the data generated, the next step is to extract the basis from the set of snapshots. In general, we hope to be able to approximate solutions within the parameter space, so, for analysis of this approach, we hold out a set of Mach cases as our test set. An ablation study of various number of training data points considered will be presented at the end of this section. But unless specifically noted, the results presented will use a dataset of 20 Mach numbers with just a single held-out solution. This is done to illustrate the best case scenarios for how well this approach applies.

\begin{figure}
    \centering
    \includegraphics[width=.7\linewidth]{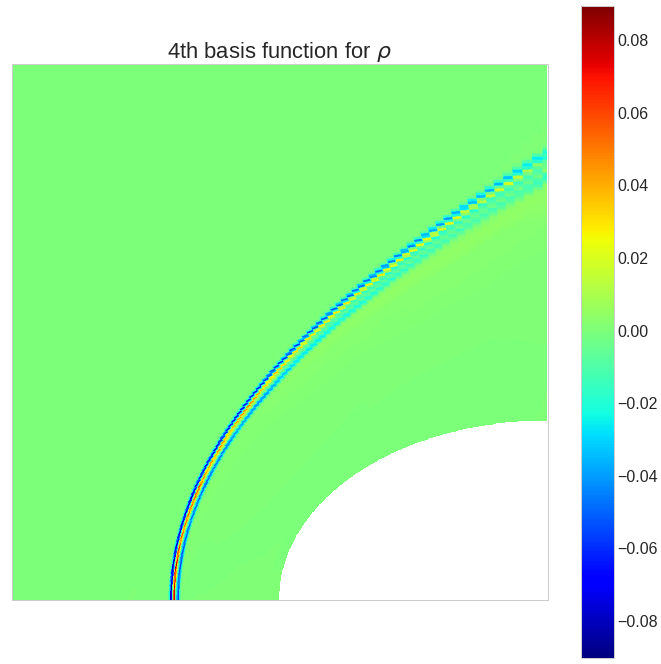}
    \caption{To demonstrate the complex spatial features inherent in higher-frequency basis functions, we visualize the predicted fourth basis for the density $\rho$ as a function of spatial coordinates $\mathbf{x}$. This is a zoomed in version around the shock region.}
    \label{fig:rho_mode3}
\end{figure}

Given the training data considered, we extract the state variables of interest and subtract the mean state from each of the solutions. Empirically, we have found that this tends to improve the speed of convergence when regressing the basis networks. The basis networks we consider are general fully-connected networks with five layers and 40 nodes per layer. We use leaky rectifed linear units~\cite{Maas13rectifiernonlinearities} as activation functions (with a negative slope of 0.01) on the hidden layers, which tends to help resolve the small-scale features in the basis functions. For example in \cref{fig:rho_mode3}, the shock is very narrow in space and changes rather rapidly. We use the Adam optimizer~\cite{Kingma2015adam} and 250 total training epochs with an initial learning rate of $10^{-3}$ while applying an exponential learning rate decay factor (0.9) after 70 training epochs.

% \dave{TODO: Consider moving the active learning equation here}
% \alex{this is an exponential learning rate decay, right?}

% In addition to the activation function we also found employing the active learning posed in equation \ref{eqn:active_learning} helps to resolve the more nuanced features.

For PDEs with complicated phenomena (shockwaves), we have found accurately representing the high-frequency basis functions can be challenging. To resolve this, we use an importance sampling technique during training that prioritizes diverse data points in the training routine. We can describe the probability $p_{i,j,l}$ of selecting a single training data point when minimizing~\cref{eqn:basis_min}, $(\mathbf{x}_l$, $U_{i,j}(\mathbf{x}_l))$, as:

\begin{equation}
    p_{i,j,l} = \frac{p^*_{i,j,l}}{\sum_{l'} p^*_{i,j,l'}},
    \label{eqn:active_learning}
\end{equation}

where $p^*_{i,j,l} = |U_{i,j}(\mathbf{x}_l) - \bar U_{i,j}(\mathbf{x}_l)|$, and $\bar U$ is the mean basis value for all spatial points in the domain $\mathbf{x}$. In practice, we have found that overusing this sampling approach leads to poor generalization. However, applying this methodology every so often (specifically every 4 epochs) improves the resolution of nuanced features.

% \alex{$p^*$ should be indexed by $i$ and $j$ still, right? then was the sum over $i$ and $j$ or just one?}
% \dave{All of them, plus the discrete $\mathbf{x}_l$ locations}
% \dave{cleaned up notation in previous eqn to reflect this as well}

% \alex{can we be a bit more explicit about how the $p_{i,j,l}$ terms are used in training? maybe say that this is being used to sample $\mathbf{x}_l$ one option would be to replace the $\sum_l$ in~\cref{eqn:basis_min} with something like $\sum_{\mathbf{x} \sim \mathbb{P}_{i,j,l}}$, where $\mathbb{P}_{i,j,l}$ is defined down here? also not sure if we can point to any literature defining similar selection criteria}
% \dave{TODO: look for literature that does this}

Once the basis networks have been sufficiently trained, the next step is to learn the set of unknown networks. For unknown network configurations, we use a seven layer feed forward network with 120 nodes per layer with leaky ReLU activation functions, similar to the basis networks. In practice, we have found that~\cref{eqn:unknowns_min} is non-trivial to learn with standard weight initialization techniques. To aid convergence of this loss term, we use a pre-training approach that attempts to regress the $C_{i,j}$ terms to known solution data to get the unknown network weights in a sufficient neighborhood before applying the operator loss in~\cref{eqn:unknowns_min}. This pre-training minimization loss can be written as:

\begin{equation}
    \min_{\Tilde C_{i,j}} \sum _{\psi} \sum_l |C_{i,j}(\mathbf{x}_l,\psi) - \Tilde C_{i,j}(\mathbf{x}_l, \psi)|^2,
    \label{eqn:pretrain_min}
\end{equation}

where $\Tilde C_{i,j}$ can be calculated through standard minimization according to the loss function:

\begin{equation}
    \min_{\tilde{C}_{i,j}} \sum _{\psi} \sum_l \left|\sum_j \tilde{C}_{i,j}(\mathbf{x}_l, \psi) \phi(\mathbf{x}_l) - \mathbf{w} (\mathbf{x}_l, \psi)\right|^2
    \label{eqn:pretrain_min2}
\end{equation}
this allows the unknown network to start off with a reasonable estimate of the desired states. It is important to note that, during this learning process, we place no dependencies on the operator portion, $\mathcal{S}_\psi\hat{\mathbf{w}}(\mathbf{x},\psi)$, in the loss function. 

After pre-training, the final step is to optimize the unknown network with respect to the governing equations, boundary conditions and any other constraints. Minimization of the governing equations and boundary conditions follows the definition of the Euler equations and the loss function defined in~\cref{eqn:unknowns_min}. For this problem we add two additional constraints to minimize the mean square error between predicted and CFD Pressures and Temperatures. Unfortunately incorporating this constraint restricts our loss function to only use $\psi$ parameters that exist in our training set. However, we have found that without incorporating this constraint leads to negative values of pressure/temperature which for this problem are non-physical.

% \dave{TODO: capitaliza Mach}

\begin{figure}[h]
    \centering
    \includegraphics[width=.7\linewidth]{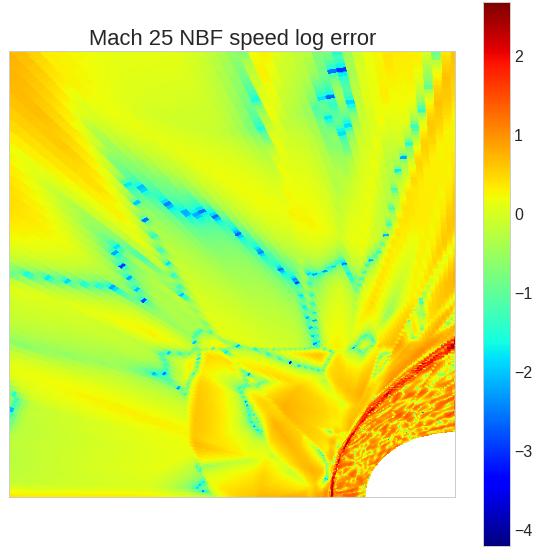}
    \caption{We evaluate our NBF prediction of speed for the Mach 25 test case with the log absolute error compared to the ground truth. The errors are highest near the boundary of the shockwave.}
    \label{fig:nbf_mach20_speed_error}
\end{figure}

Finally with the basis and unknown networks trained, we are able to infer new solutions within the defined parameter space. We seek to generalize performance across the parameter space $\Psi$ of Mach numbers. \Cref{fig:nbf_mach20_speed} shows the NBF predicted solution for the speed within the domain while considering a Mach 25 inflow condition. \Cref{fig:nbf_mach20_speed_error} shows the log absolute error of this predicted solution with respect to the CFD++ data. It is worth noting that this particular scenario is held out of our training data set and still shows good predictive performance.

\begin{figure}[h]
    \centering
    \includegraphics[width=\linewidth]{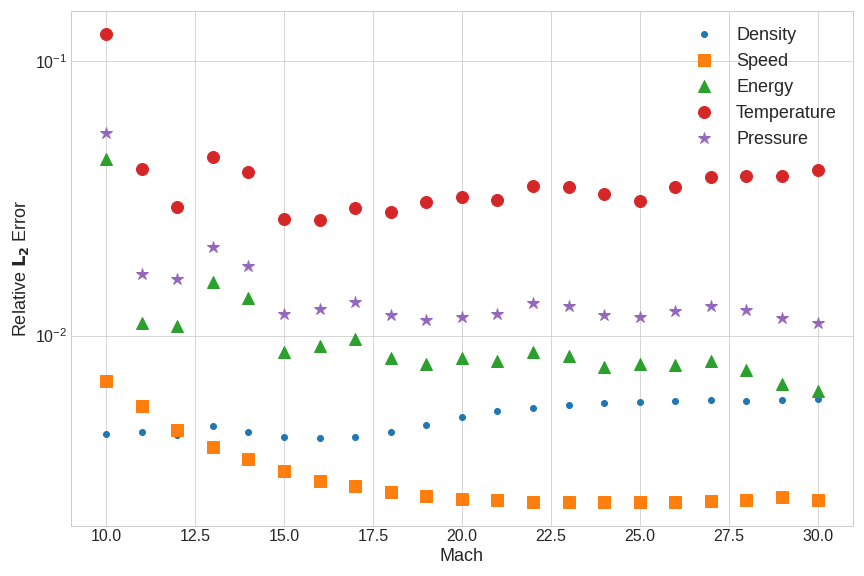}
    \caption{Relative $L_2$ error for variables of interest with respect to Mach number, for the test data (Mach 25) and training data (remainder).}
    \label{fig:nbf_mach_errors}
\end{figure}

It is also useful to understand how well the NBF approximation works across the parameter space. \Cref{fig:nbf_mach_errors} shows the relative $L_2$ error as a function of the Mach number. The errors appear to be relatively stable across the parameter space with slightly better performance for the higher Mach numbers.

\subsection{Comparison with Vanilla DeepONet}\label{sec:DeepONet}
In order to understand how well the NBF approach compares against existing approaches, we will benchmark our solution against the Vanilla DeepONet operator approach. The equation for the vanilla DeepONet is given by~\cite{lu2022comprehensive}:

\begin{equation}
    \mathcal{G}(v)(\xi) = \sum _{j=1}^q b_j(v) t_j(\xi) + b_0
    \label{eqn:vanillaDeepONet}
\end{equation}
where $v$ and $\xi$ are the parameter (in our case $\psi$) and spatial values (in our case $\mathbf{x}$). \cite{lu2022comprehensive} focuses on the case where  $v$ is a function that parameterizes the PDE -- for example, an initial condition. For comparison purposes, we will consider it a parameterization of $\Psi$. Additionally, $b_j(v)$ and $t_j(\xi)$ represent the branch and trunk networks respectively, with $b_0$ acting as a bias imposed on the outputs.

Since we are interested in the two dimensional steady state Euler equations, we require a DeepONet representation of multiple state variables defined in ~\cref{eq:euler_components}. Making use of recommendations from \cite{lu2022comprehensive}, we implement a set of feed-forward DeepONets for each state variable considered. The network architectures that we found generated the best performance for our use case comprised of: 5 layers of 120 nodes with hyperbolic tangent activation functions for the branch networks and 6 layers of 120 nodes with rectified linear units for the trunk networks. These networks were then connected in a shared layer of 64 nodes combined with a bias ($b_0$) to produce predictions for each state variable. During training, we used the same Adam optimizer, learning rate decay and number of epochs as was used for the NBF approach.

We added a few components that appeared to aid the training process for our problem specifically: adding prioritized spatial sampling and zero-mean unit-variance (ZMUV) normalization functions for spatial/parameter inputs as well as the output state variables. For spatial sampling prioritization we used the \cref{eqn:active_learning} to more often select training examples with more deviation from the mean solution. Additionally, adding the ZMUV normalization to input/output parameters appeared to speed convergence and prevent collapsing the solution to the mean flow.

\begin{figure}[h]
    \centering
    \includegraphics[width=\linewidth]{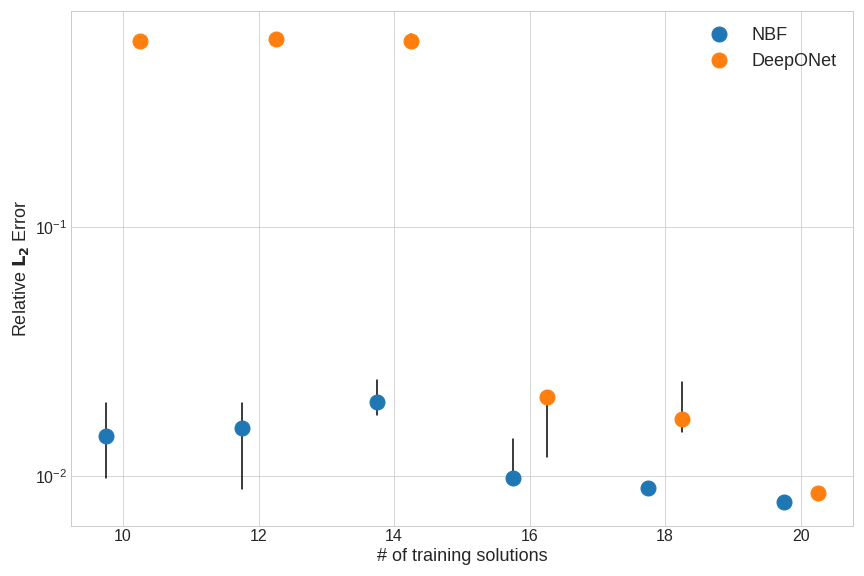}
    \caption{As an ablation study, we vary the amount of training data and evaluate $L_2$ relative error for the test set. The dots represent the median error with the black tails indicating inner 50th percentile. Blue show NBF relative error compared to the orange of the vanilla DeepONet}
    \label{fig:ablation_study}
\end{figure}

\cref{fig:ablation_study} shows the comparison for the vanilla DeepONet vs the Neural Basis Function approach considering the total number of training solutions as inputs. We are able to show better performance in the approximation across the number of training examples. Specifically we are able to produce significantly better results for smaller training data sets, indicating better generalization for sparse data. We believe we are able to show this better performance due to the pre-conditioned nature of basis functions included in the NBF estimates as well as application of the governing equation as regularization during training.

It is worth noting, that although the training process lines up reasonably closely between the two approaches, there is some overhead incurred by the NBF approach in: calculating the SVD, then training the unknown network. Additionally, the DeepONet architecture is significantly smaller than the NBF architecture as it is described here. We believe there are architecture optimizations that could reduce the size and will consider this for future work. Thus, it is not a completely direct comparison but is as close as we could get to provide a baseline for the NBF approach.

We found that for many of the cases within the ablation study, the Energy ($E$) state variable was the most difficult to get to converge. This is likely due to the magnitude of values compared to the other state variables ($\rho$, $u$ and $v$). For DeepONet, applying the ZMUV transformation on the output state improved the convergence but not significantly so.

\subsection{Acceleration of CFD solutions}\label{sec:acceleration}

% \dave{What are the parameters that we used to initialize the solutions?}
% \dave{What is the RHS Average value a measure of? is it $\frac{\partial F}{partial t}$?}
% \dave{Are there additional details that go into the acceleration process worth mentioning?}

In the conventional CFD approach an initial condition is employed with iterative updates until steady state is reached. If one was to use an initial condition that is close to the converged solution, fewer iterations would be required to reach the high-fidelity steady-state. The NBF solution can be leveraged to accelerate the CFD solver by providing this nearly converged initial condition. In the following section, we will compare the performance of the conventional CFD approach with NBF informed CFD approach. The two approaches will be evaluated by comparing their average residual history. The time-dependent form of the Euler equation includes the the term:

\begin{equation}
    \frac{\partial \mathbf{w}}{\partial t}
\end{equation}
and adds to \cref{eq:euler_components}. Additional details on the setup and implementation of the CFD++ configuration can be found in~\cref{sec:cfd++_details}
 
The convergence is determined by the evolution of the time-dependent residual. Once the residual levels off, the simulation is deemed to be converged and the solution has reached steady-state. In the CFD++ finite volume approach, the governing equations are integrated over the cell volumes. For the case of the Euler equations, the right hand side becomes the sum of the scalar product of the inviscid fluxes and the cell face areas.  For a given equation, the residual is defined as the absolute value of the residual divided by the cell volume averaged over all the cells. Since we are working with a system of equations, we monitor the average residual of our system equations, i.e. the sum of the residual of each equation divided by the number of equations. Additional detail of all the numerical procedures can be found in the CFD++ user-manual\footnote{https://www.metacomptech.com/index.php/features/icfd}.

\begin{figure}[h]
    \centering
    \includegraphics[width=\linewidth]{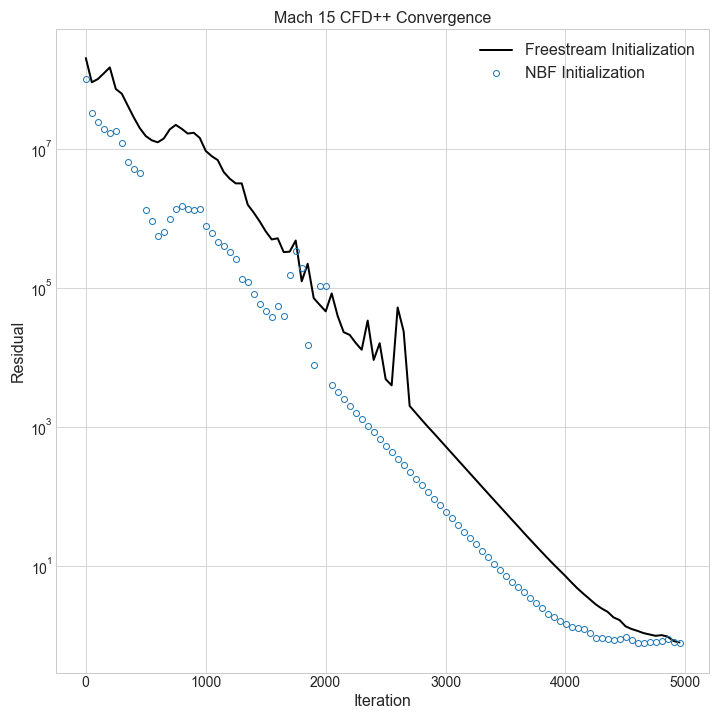}
    \label{fig:nbf_mach15_convergence}
    \caption{Steady state convergence acceleration of a Mach 15 scenario. Using the NBF prediction as the initialization results in a speedup of roughly 750 iterations compared to the freestream initialization.}
\end{figure}

\begin{figure}[h]
    \centering
    \includegraphics[width=\linewidth]{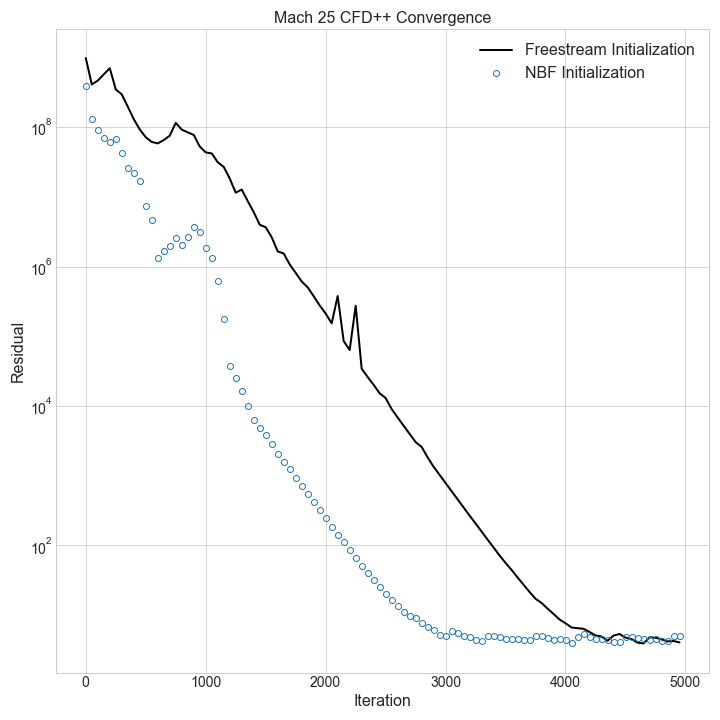}
    \label{fig:nbf_mach25_convergence}
    \caption{Steady state convergence acceleration of a Mach 25 scenario. Using the NBF prediction as the initialization results in a speedup of roughly 1000 iterations, compared to the freestream initialization.}
\end{figure}

\Cref{fig:nbf_mach15_convergence} and \cref{fig:nbf_mach25_convergence} show the time-dependent residual as a function of the number of iterations for Mach 15 and 25 respectively. As we can see from these plots, the NBF solution provides a noticeable speed up in the convergence of the high fidelity solver.

\section{Discussion}\label{sec:discussion}

This work has shown the potential for using an operator learning based approach, like the Neural Basis Function, as an acceleration mechanism for high-fidelity CFD solvers. We found that regressing a set of networks to POD basis data, then using an additional unknown network in linear combination can provide sufficient estimates of state variables within a configurable parameter space. High Mach flow cases often induce complicated phenomena, like discontinuous shockwaves, which are difficult to represent numerically, even with standard techniques. Using the 2D steady state Euler equations as an example, we were able to produce accurate estimates to the flow field for high Mach (10-30) scenarios. These solutions were then used as initial conditions to a high-fidelity CFD solver (CFD++) to improve the convergence, resulting in speed ups up to 1000 iterations.

As future work on the application side, we intend to extend these results to the full Navier-Stokes equations, while transitioning to a three-dimensional domain. Along the way improvements will be made to the methodology to better understand network architecture performance as well as improving training techniques for the basis and unknown networks. We also hope to develop a methodology using NBF (or other operator approaches) to handle problems where the spatial domain might vary across snapshots. This would allow for the ability to do shape optimization or analysis of multiple geometric configurations.

On the learning side, we hope to factor in an active learning feedback loop \cite{settles2009active} that will be able to identify regions of the $\psi$ parameter space with poor predictive performance, in order to execute more high-fidelity scenarios in these regions to improve overall accuracy. Finally, some combination of DeepOnet and NBF seems like a reasonable path forward especially for speeding convergence of these techniques. For example, one could create a set of trunk networks that are regressed from the orthogonal SVD basis, then freeze their weights and biases. Then adding a vanilla DeepONet in combination with these bases could greatly speed the convergence and make it easier to implement an active learning type of approach for data generation.

% % Acknowledgements should only appear in the accepted version.
% \section*{Acknowledgements}

% \textbf{Do not} include acknowledgements in the initial version of
% the paper submitted for blind review.

% If a paper is accepted, the final camera-ready version can (and
% probably should) include acknowledgements. In this case, please
% place such acknowledgements in an unnumbered section at the
% end of the paper. Typically, this will include thanks to reviewers
% who gave useful comments, to colleagues who contributed to the ideas,
% and to funding agencies and corporate sponsors that provided financial

\bibliography{references}

\begin{thebibliography}{27}
\providecommand{\natexlab}[1]{#1}
\providecommand{\url}[1]{\texttt{#1}}
\expandafter\ifx\csname urlstyle\endcsname\relax
  \providecommand{\doi}[1]{doi: #1}\else
  \providecommand{\doi}{doi: \begingroup \urlstyle{rm}\Url}\fi

\bibitem[Baydin et~al.(2017)Baydin, Pearlmutter, Radul, and
  Siskind]{Baydin2017autodiff}
Baydin, A.~G., Pearlmutter, B.~A., Radul, A.~A., and Siskind, J.~M.
\newblock Automatic differentiation in machine learning: A survey.
\newblock \emph{J. Mach. Learn. Res.}, 18\penalty0 (1):\penalty0 5595–5637,
  jan 2017.
\newblock ISSN 1532-4435.

\bibitem[Berkooz et~al.(1993)Berkooz, Holmes, and Lumley]{berkooz1993proper}
Berkooz, G., Holmes, P., and Lumley, J.~L.
\newblock The proper orthogonal decomposition in the analysis of turbulent
  flows.
\newblock \emph{Annual review of fluid mechanics}, 25\penalty0 (1):\penalty0
  539--575, 1993.

\bibitem[Bhattacharya et~al.(2021)Bhattacharya, Hosseini, Kovachki, and
  Stuart]{Bhattacharya2021pdes}
Bhattacharya, K., Hosseini, B., Kovachki, N.~B., and Stuart, A.~M.
\newblock Model {Reduction} {And} {Neural} {Networks} {For} {Parametric}
  {PDEs}.
\newblock \emph{The SMAI journal of computational mathematics}, 7:\penalty0
  121--157, 2021.
\newblock \doi{10.5802/smai-jcm.74}.

\bibitem[Certik(2017)]{certik2017theoretical}
Certik, O.
\newblock Theoretical physics reference.
\newblock 2017.

\bibitem[Gunzburger et~al.(2007)Gunzburger, Peterson, and
  Shadid]{gunzburger2007reduced}
Gunzburger, M.~D., Peterson, J.~S., and Shadid, J.~N.
\newblock Reduced-order modeling of time-dependent pdes with multiple
  parameters in the boundary data.
\newblock \emph{Computer methods in applied mechanics and engineering},
  196\penalty0 (4-6):\penalty0 1030--1047, 2007.

\bibitem[Han et~al.(2018)Han, Jentzen, and E]{Han2018PDEs}
Han, J., Jentzen, A., and E, W.
\newblock Solving high-dimensional partial differential equations using deep
  learning.
\newblock \emph{Proceedings of the National Academy of Sciences}, 115\penalty0
  (34):\penalty0 8505--8510, 2018.
\newblock ISSN 0027-8424.
\newblock \doi{10.1073/pnas.1718942115}.

\bibitem[He \& Zou(2021)He and Zou]{functorch}
He, H. and Zou, R.
\newblock functorch: Jax-like composable function transforms for pytorch.
\newblock \url{https://github.com/pytorch/functorch}, 2021.

\bibitem[Karniadakis et~al.(2021)Karniadakis, Kevrekidis, Lu, Perdikaris, Wang,
  and Yang]{karniadakis2021physics}
Karniadakis, G.~E., Kevrekidis, I.~G., Lu, L., Perdikaris, P., Wang, S., and
  Yang, L.
\newblock Physics-informed machine learning.
\newblock \emph{Nature Reviews Physics}, 3\penalty0 (6):\penalty0 422--440,
  2021.

\bibitem[Kashinath et~al.(2021)Kashinath, Mustafa, Albert, Wu, Jiang,
  et~al.]{Kashinath2021ml_climate}
Kashinath, K., Mustafa, M., Albert, A., Wu, J.-L., Jiang, C., et~al.
\newblock Physics-informed machine learning: case studies for weather and
  climate modelling.
\newblock \emph{Philosophical Transactions of the Royal Society A:
  Mathematical, Physical and Engineering Sciences}, 379\penalty0
  (2194):\penalty0 20200093, 2021.
\newblock \doi{10.1098/rsta.2020.0093}.

\bibitem[Khan \& Lowther(2022)Khan and Lowther]{Khan2022e&m}
Khan, A. and Lowther, D.~A.
\newblock Physics informed neural networks for electromagnetic analysis.
\newblock \emph{IEEE Transactions on Magnetics}, pp.\  1--1, 2022.
\newblock \doi{10.1109/TMAG.2022.3161814}.

\bibitem[Kingma \& Ba(2015)Kingma and Ba]{Kingma2015adam}
Kingma, D.~P. and Ba, J.
\newblock Adam: {A} method for stochastic optimization.
\newblock In Bengio, Y. and LeCun, Y. (eds.), \emph{3rd International
  Conference on Learning Representations, {ICLR} 2015, San Diego, CA, USA, May
  7-9, 2015, Conference Track Proceedings}, 2015.

\bibitem[Li et~al.(2020)Li, Kovachki, Azizzadenesheli, Liu, Bhattacharya,
  Stuart, and Anandkumar]{li2020fourier}
Li, Z., Kovachki, N., Azizzadenesheli, K., Liu, B., Bhattacharya, K., Stuart,
  A., and Anandkumar, A.
\newblock Fourier neural operator for parametric partial differential
  equations.
\newblock \emph{arXiv preprint arXiv:2010.08895}, 2020.

\bibitem[Lu et~al.(2019{\natexlab{a}})Lu, Jin, and Karniadakis]{lu2019deeponet}
Lu, L., Jin, P., and Karniadakis, G.~E.
\newblock Deeponet: Learning nonlinear operators for identifying differential
  equations based on the universal approximation theorem of operators.
\newblock \emph{arXiv preprint arXiv:1910.03193}, 2019{\natexlab{a}}.

\bibitem[Lu et~al.(2019{\natexlab{b}})Lu, Meng, Mao, and
  Karniadakis]{Lu2019deepxde}
Lu, L., Meng, X., Mao, Z., and Karniadakis, G.~E.
\newblock Deepxde: {A} deep learning library for solving differential
  equations.
\newblock \emph{CoRR}, abs/1907.04502, 2019{\natexlab{b}}.

\bibitem[Lu et~al.(2022)Lu, Meng, Cai, Mao, Goswami, Zhang, and
  Karniadakis]{lu2022comprehensive}
Lu, L., Meng, X., Cai, S., Mao, Z., Goswami, S., Zhang, Z., and Karniadakis,
  G.~E.
\newblock A comprehensive and fair comparison of two neural operators (with
  practical extensions) based on fair data.
\newblock \emph{Computer Methods in Applied Mechanics and Engineering},
  393:\penalty0 114778, 2022.

\bibitem[Maas et~al.(2013)Maas, Hannun, and Ng]{Maas13rectifiernonlinearities}
Maas, A.~L., Hannun, A.~Y., and Ng, A.~Y.
\newblock Rectifier nonlinearities improve neural network acoustic models.
\newblock In \emph{in ICML Workshop on Deep Learning for Audio, Speech and
  Language Processing}, 2013.

\bibitem[Mao et~al.(2020)Mao, Jagtap, and Karniadakis]{mao2020shockwavepinn}
Mao, Z., Jagtap, A.~D., and Karniadakis, G.~E.
\newblock Physics-informed neural networks for high-speed flows.
\newblock \emph{Computer Methods in Applied Mechanics and Engineering},
  360:\penalty0 112789, 2020.
\newblock ISSN 0045-7825.
\newblock \doi{https://doi.org/10.1016/j.cma.2019.112789}.

\bibitem[Mao et~al.(2021)Mao, Lu, Marxen, Zaki, and
  Karniadakis]{mao2021deepmmnet}
Mao, Z., Lu, L., Marxen, O., Zaki, T.~A., and Karniadakis, G.~E.
\newblock Deepm\&mnet for hypersonics: Predicting the coupled flow and
  finite-rate chemistry behind a normal shock using neural-network
  approximation of operators.
\newblock \emph{Journal of Computational Physics}, 447:\penalty0 110698, 2021.
\newblock ISSN 0021-9991.
\newblock \doi{https://doi.org/10.1016/j.jcp.2021.110698}.

\bibitem[Paszke et~al.(2019)Paszke, Gross, Massa, Lerer, Bradbury,
  et~al.]{pytorch}
Paszke, A., Gross, S., Massa, F., Lerer, A., Bradbury, J., et~al.
\newblock Pytorch: An imperative style, high-performance deep learning library.
\newblock In Wallach, H., Larochelle, H., Beygelzimer, A., d\textquotesingle
  Alch\'{e}-Buc, F., Fox, E., and Garnett, R. (eds.), \emph{Advances in Neural
  Information Processing Systems 32}, pp.\  8024--8035. Curran Associates,
  Inc., 2019.

\bibitem[Quarteroni et~al.(2015)Quarteroni, Manzoni, and
  Negri]{quarteroni2015reduced}
Quarteroni, A., Manzoni, A., and Negri, F.
\newblock \emph{Reduced basis methods for partial differential equations: an
  introduction}, volume~92.
\newblock Springer, 2015.

\bibitem[Raissi et~al.(2019)Raissi, Perdikaris, and
  Karniadakis]{raissi2019physics}
Raissi, M., Perdikaris, P., and Karniadakis, G.~E.
\newblock Physics-informed neural networks: A deep learning framework for
  solving forward and inverse problems involving nonlinear partial differential
  equations.
\newblock \emph{Journal of Computational Physics}, 378:\penalty0 686--707,
  2019.

\bibitem[Settles(2009)]{settles2009active}
Settles, B.
\newblock Active learning literature survey.
\newblock 2009.

\bibitem[Sirovich(1987)]{sirovich1987turbulence}
Sirovich, L.
\newblock Turbulence and the dynamics of coherent structures. i. coherent
  structures.
\newblock \emph{Quarterly of applied mathematics}, 45\penalty0 (3):\penalty0
  561--571, 1987.

\bibitem[Willard et~al.(2020)Willard, Jia, Xu, Steinbach, and
  Kumar]{Willard2020physicsai}
Willard, J., Jia, X., Xu, S., Steinbach, M., and Kumar, V.
\newblock Integrating scientific knowledge with machine learning for
  engineering and environmental systems, 2020.
\newblock URL \url{https://arxiv.org/abs/2003.04919}.

\bibitem[Willcox \& Peraire(2002)Willcox and Peraire]{willcox2002balanced}
Willcox, K. and Peraire, J.
\newblock Balanced model reduction via the proper orthogonal decomposition.
\newblock \emph{AIAA journal}, 40\penalty0 (11):\penalty0 2323--2330, 2002.

\bibitem[Witman et~al.(2017)Witman, Gunzburger, and
  Peterson]{witman2017reduced}
Witman, D.~R., Gunzburger, M., and Peterson, J.
\newblock Reduced-order modeling for nonlocal diffusion problems.
\newblock \emph{International Journal for Numerical Methods in Fluids},
  83\penalty0 (3):\penalty0 307--327, 2017.

\bibitem[Yazdani et~al.(2021)Yazdani, Deng, Li, Javadi, Li, Jamali, Lin,
  Humphrey, Mantzoros, and Em~Karniadakis]{Yazdani2021bloodcell}
Yazdani, A., Deng, Y., Li, H., Javadi, E., Li, Z., Jamali, S., Lin, C.,
  Humphrey, J.~D., Mantzoros, C.~S., and Em~Karniadakis, G.
\newblock Integrating blood cell mechanics, platelet adhesive dynamics and
  coagulation cascade for modelling thrombus formation in normal and diabetic
  blood.
\newblock \emph{Journal of The Royal Society Interface}, 18\penalty0
  (175):\penalty0 20200834, 2021.
\newblock \doi{10.1098/rsif.2020.0834}.

\end{thebibliography}
\bibliographystyle{icml2022}

%%%%%%%%%%%%%%%%%%%%%%%%%%%%%%%%%%%%%%%%%%%%%%%%%%%%%%%%%%%%%%%%%%%%%%%%%%%%%%%
%%%%%%%%%%%%%%%%%%%%%%%%%%%%%%%%%%%%%%%%%%%%%%%%%%%%%%%%%%%%%%%%%%%%%%%%%%%%%%%
% APPENDIX
%%%%%%%%%%%%%%%%%%%%%%%%%%%%%%%%%%%%%%%%%%%%%%%%%%%%%%%%%%%%%%%%%%%%%%%%%%%%%%%
%%%%%%%%%%%%%%%%%%%%%%%%%%%%%%%%%%%%%%%%%%%%%%%%%%%%%%%%%%%%%%%%%%%%%%%%%%%%%%%
\newpage
\appendix
\onecolumn
\section{Euler Equation Expansion}\label{sec:appendix_euler_expansion}

We can expand the Euler equations in terms of our state variables $\mathbf{w}$:

\begin{eqnarray*}
    \mathbf{F}_1    &=& [f_{11}, f_{12}, f_{13}, f_{14}]^T\\
    f_{11} &=& \rho u\\
           &=& w_0 w_1\\
    f_{12} &=& \rho u^2 + p\\
           &=& w_0 w_1^2 + (\gamma - 1)[w_3 - \frac{1}{2}w_0(w_1^2 + w_2^2)]\\
    f_{13} &=& \rho u v\\
           &=& w_0 w_1 w_2\\
    f_{14} &=& (E+p) u\\
           &=& (w_3 + (\gamma - 1)[w_3 - \frac{1}{2}w_0(w_1^2 + w_2^2)]) w_1\\
    \mathbf{F}_2    &=& [f_{21}, f_{22}, f_{23}, f_{24}]^T\\
    f_{21} &=& \rho v\\
           &=& w_0 w_2\\
    f_{22} &=& \rho u v\\
           &=& w_0 w_1 w_2\\
    f_{23} &=& \rho v^2 + p\\
           &=& w_0 w_2^2 + (\gamma - 1)[w_3 - \frac{1}{2}w_0(w_1^2 + w_2^2)]\\
    f_{24} &=& (E+p)v\\
           &=& (w_3 + (\gamma - 1)[w_3 - \frac{1}{2}w_0(w_1^2 + w_2^2)])w_2
\end{eqnarray*}

\twocolumn

\section{Other Variable Field Visuals}\label{sec:variable_fields}
The following set of plots provide a visual for the other variables of interest using the NBF approximation to the Mach 25 scenario.

\begin{figure}[h!]
    \centering
    \includegraphics[width=.6\linewidth]{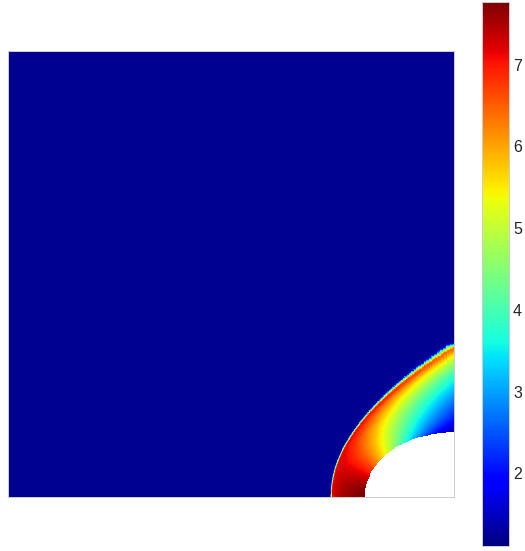}
    \caption{Density ($\rho$)}
\end{figure}

\begin{figure}[h!]
    \centering
    \includegraphics[width=.6\linewidth]{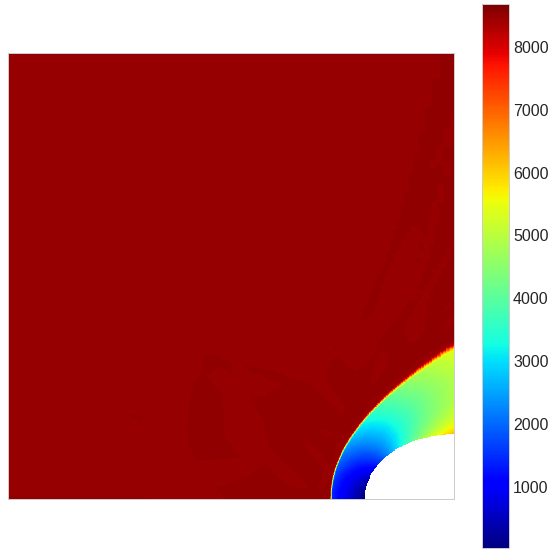}
    \caption{$x$-velocity component ($u$)}
\end{figure}

\begin{figure}[h!]
    \centering
    \includegraphics[width=.6\linewidth]{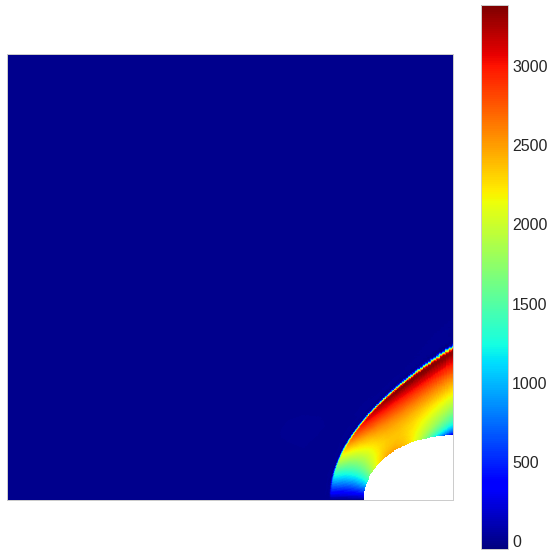}
    \caption{$y$-velocity component ($v$)}
\end{figure}

\begin{figure}[h!]
    \centering
    \includegraphics[width=.6\linewidth]{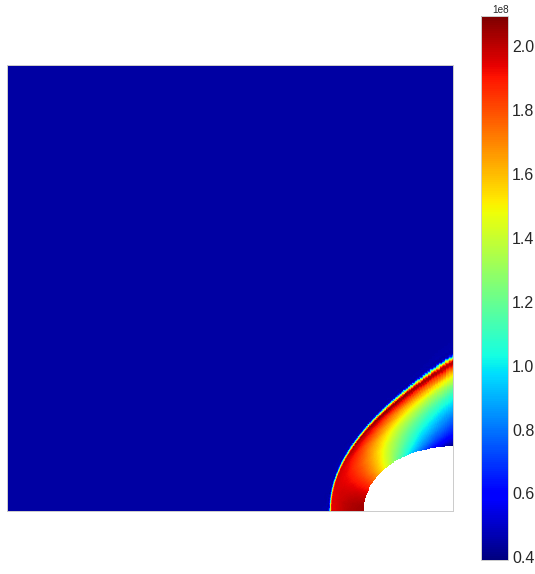}
    \caption{Energy ($E$)}
\end{figure}

\begin{figure}[h!]
    \centering
    \includegraphics[width=.6\linewidth]{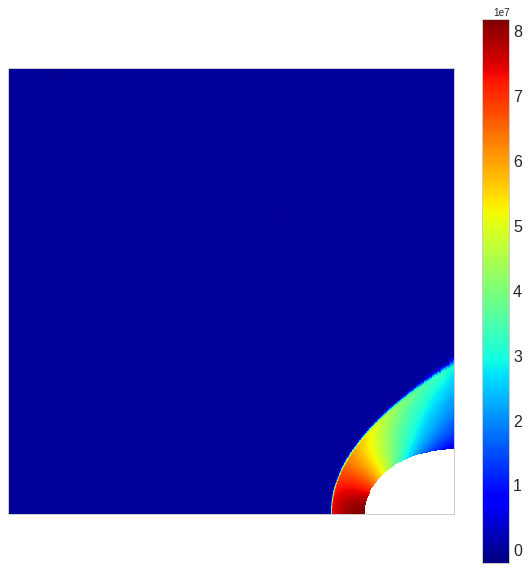}
    \caption{Pressure ($P$)}
\end{figure}

\begin{figure}[h!]
    \centering
    \includegraphics[width=.6\linewidth]{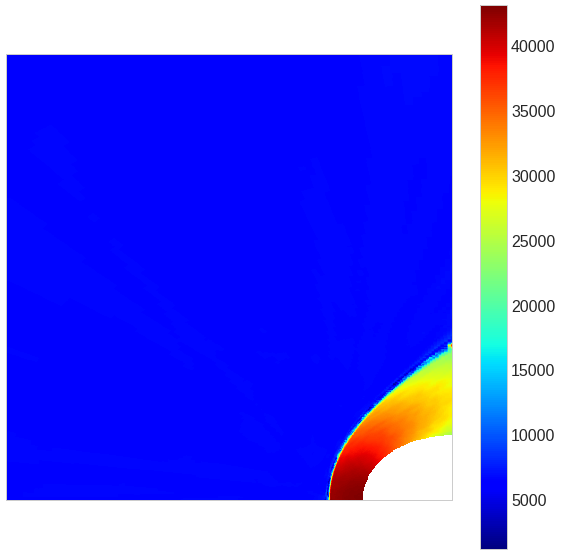}
    \caption{Temperature ($T$)}
\end{figure}

\onecolumn
\section{CFD++ Details}\label{sec:cfd++_details}

In our investigation, CFD++ was run with a second order spatial discretization scheme. The inviscid fluxes were calculated based on the minmod TVD limiter. The fluxes on the cell faces were reconstructed based on centroidal polynomials. Spatial scheme blending was leveraged for the sake of numerical stability. During the initial 250 iterations, first order fluxes were used exclusively. Then a linear blending of the first and second order discretization scheme was used until iteration 750, beyond which the second order fluxes were used exclusively.

The time-residual from iteration to iteration depends on the integration scheme. In this study, an implicit Gauss-Seidel relaxation time integration scheme was utilized with no additional convergence acceleration methods.  The time step sizes were based on the local (spatially varying) CFL (Courant Friedrichs-Lewy) number. The CFL number is a dimensionless number that is equal to the time step size scaled by the time it takes for the fasts moving characteristics to travel across a cell $CFL= \frac{dt}{\frac{dx}{|\mathbf{u}|+a}}$, recalling $|\mathbf{u}|$ is the scaler speed field.  $a$ is the speed of sound ($a = \sqrt{T*R*\gamma}$) In each simulations, the CFL number was linearly increased from 0.01 to 10 over the first 1000 iterations beyond which the CFL number was kept constant. 

%%%%%%%%%%%%%%%%%%%%%%%%%%%%%%%%%%%%%%%%%%%%%%%%%%%%%%%%%%%%%%%%%%%%%%%%%%%%%%%
%%%%%%%%%%%%%%%%%%%%%%%%%%%%%%%%%%%%%%%%%%%%%%%%%%%%%%%%%%%%%%%%%%%%%%%%%%%%%%%

\end{document}